\documentclass{article}
\usepackage{iclr2027_conference,times}

\usepackage{amsmath}
\usepackage{booktabs}
\usepackage{graphicx}
\usepackage{algorithm}
\usepackage{algorithmic}
\usepackage{xspace}
\usepackage{xcolor}

\usepackage{mathtools}
\usepackage{subcaption}
\usepackage{placeins}
\usepackage{flafter}
\usepackage[hidelinks]{hyperref}
\usepackage[utf8]{inputenc}
\usepackage[T1]{fontenc}
\usepackage{newtxtext,newtxmath}

\mathtoolsset{showonlyrefs=true}

\newcommand{\RMSNorm}{\operatorname{RMSNorm}}
\newcommand{\ATTN}{\operatorname{ATTN}}
\newcommand{\FFN}{\operatorname{FFN}}
\newcommand{\embed}{\operatorname{embed}}
\newcommand{\unembed}{\operatorname{unembed}}
\newcommand{\Last}{\operatorname{Last}}

\title{The Extender: A Log-Structured Transformer}

\author{Jakob Eriksson \\
  University of Illinois Chicago \\
  \texttt{jakob@uic.edu}}
\date{July 21, 2026}
\hypersetup{pdftitle={The Extender: A Log-Structured Transformer},pdfauthor={Jakob Eriksson}}

\newcommand{\xstar}{$\mathbf{x}_\ast$\xspace}

\newcommand{\xstarcache}{\xstar-cache\xspace}
\newcommand{\kvcache}{$\mathbf{kv}$-cache\xspace}

\begin{document}
\maketitle

\begin{abstract}
We introduce the Extender, a log-structured variant of the standard Transformer architecture.
In a standard Transformer, each layer communicates with subsequent layers exclusively via the residual $\mathbf{h}$, a superposition channel.
The Extender adds a concatenation channel $\mathbf{x}$: each layer $\ell$ emits both a residual update $\delta_\ell$ which is added to $\mathbf{h}$, and a much smaller {\it extension} $\epsilon_\ell$ which is appended to $\mathbf{x}$.
While both the FFN and $\mathbf{q}$ see $\mathbf{h}$, the attention $\mathbf{kv}$ projections take only $\mathbf{x}$ as input. As a result, the fully {\it extended} $\mathbf{x}$ contains the complete input for the $\mathbf{kv}$ projections of all layers,
reducing the persistent attention memory footprint from $2Ld_{model}$ to $\sum|\epsilon_\ell|$.
We find that with $|\epsilon_\ell|=32$, the Extender matches Transformer accuracy on short-context (CORE) tasks at 199M-924M parameters, and
exceeds Transformer accuracy on long-context (RULER) workloads, again at 924M parameters.
For our 1664-wide, 924M model, the Extender's persistent attention memory footprint is $104\times$ smaller than MHA.
The memory savings grow with model width.
\end{abstract}

\section{Introduction}

The residual stream is fundamental to the standard Transformer model: it is the sole input to every layer; the sole output of each layer is an update to the residual stream.
Arguably, its most important function is that of mitigating vanishing gradients in deep models: each individual layer sees a direct gradient from the final loss computation, enabling stable optimization.
In other words, the loss directly teaches each layer how to predict the next token.

However, this design is not without drawbacks. While each layer may in practice only make small changes, the residual update from each layer is full width, making each feature of the input to the subsequent attention layer unique.
During inference, this requires each attention layer to maintain its own, full-width residual ($\mathbf{h}$) or key-value ($\mathbf{kv}$) cache to avoid a very costly recomputation.
The memory footprint of these caches grows linearly with model width $d_{model}$, depth $L$ and context length $T$. For a large language model,
the aggregate \kvcache alone can exceed 100 GB before dimensionality reduction, layer reuse, quantization and other space-saving optimizations.

Moreover, beyond predicting the next token, each layer $\ell$ may also need to communicate its own findings, say $\phi_\ell$ to subsequent layers.
Here, $\phi_\ell$ may not look anything like the next token; for example, it may encode a distance from another token, or a repetition counter, yet this too must be communicated through the residual, potentially in contention with other writes, and against the teachings of the loss.

The {\it Extender} is a log-structured variation on the classic Transformer architecture that separates the next-token prediction stream from the attention input.
Similar to the Transformer, the Extender is made up of a stack of Extender layers, each consisting of several individual sub-layers, but dominated by softmax attention and FFN.
However, instead of giving the attention sub-layers full visibility into the output of the previous layer, the Extender provides two {\it distinct} channels:

\begin{description}
  \item[$\mathbf{h}$] The conventional residual, used primarily for predicting the next token. Each layer contributes a full-width {\it residual update} $\mathbf{\delta}_\ell$, such that $\mathbf{h}_\ell=\mathbf{h}_{\ell-1}+w_\ell \mathbf{\delta}_\ell$, where $w_\ell$ is a scalar weight. $\mathbf{h}_\ell$ is part of the input to the {\it FFN sub-layer} of the next block, as well as to attention ($\mathbf{q}$ only, not $\mathbf{kv}$), and the final logit projection. $\mathbf{h}_{-1}=[embed(t)]$.
  \item[$\mathbf{x}$] Each layer $\ell$ contributes information to subsequent attention $\mathbf{kv}$ and $\mathbf{q}$ projections via a small, read-only {\it feature extension} $\epsilon_\ell$ to a log-structured token embedding. The extended embedding after layer $\ell$ is $\mathbf{x}_\ell=[\mathbf{x}_{\ell-1};RMSNorm(\epsilon_\ell)]$ where $\mathbf{x}_{\text{-}1}=[embed(t)]$.
\end{description}

We use \xstar to refer to the {\it fully extended} embedding after the token has traversed all $L$ layers. As the $\ast$ in the notation implies, the append-only nature of $\mathbf{x}$ in the Extender architecture means that \xstar contains every $x_\ell, \ell \in [0..L\text{-}1]$ as a prefix.
Therefore, while every layer sees a different $\mathbf{x}_\ell$, a single shared \xstar-cache suffices across all attention sub-layers.

Thus while the Multi-Head Attention (MHA) \cite{vaswani2017attention} Transformer requires \kvcache memory proportional to $2TLd_{\rm model}$ (two $d_{model}$-wide vectors per layer, per token), the Extender instead requires \xstar-cache memory $TLd_{\epsilon}$ (one $d_\epsilon$-wide extension per layer, per token), reducing the persistent attention memory footprint at long context lengths by up to $\frac{2d_{model}}{d_\epsilon}\times$ vs. MHA. The model sizes we tested, 198M/436M/920M, with $d_\epsilon{=}32$, the Extender matches our Reference Transformer on short-context tasks ($\sim+1$ on CORE), and leads significantly on long-context (RULER) tasks.

To compute softmax attention without a \kvcache, \xstar-cache based attention would have to repeatedly re-materialize $\mathbf{kv}$ from \xstar, potentially resulting in significant computational overhead. To resolve this, the  Extender features an {\it ephemeral} \kvcache{} at inference time. The {\it ephemeral} \kvcache of a given layer may be materialized on-demand from the \xstar-cache by multiplication with the relevant attention $W_k$ and $W_v$ matrices. Amortized over multiple input and output tokens, this $\mathbf{kv}$ materialization cost is small. Thus the {\it ephemeral} \kvcache may be released between turns, as needed.

Because of the large attention memory footprint of MHA, it is rarely used in its pure form in large models. Instead, models that retain softmax attention use techniques such as dimensionality reduction \cite{shazeer2019mqa,ainslie2023gqa, mla}, layer reuse \cite{minicache, sun2024yoco,wu2024lckv,brandon2024cla,zuhri2024mlkv} and quantization \cite{xiao2023smoothquant,frantar2023gptq,liu2024kivi} to limit the footprint.
The Extender instead changes the architecture, separating the per-layer next-token prediction $\delta_\ell$ from the per-token features $\epsilon_\ell$ that persist in attention memory.
The result is a compact attention memory footprint between turns, while achieving similar accuracy on short-context and better accuracy on long-context tasks.
During turns, memory requirements are unchanged vs a standard Transformer. However, existing dimensionality reduction, layer reuse and quantization methods may still be applied to attention during turns, if desired.
Our experiments with GQA (See Appendix \ref{s:gqa}) suggest that Grouped Query Attention combines well with the Extender.

The remainder of the paper is organized as follows. \S\ref{s:related} briefly introduces past work on reducing attention memory footprint. \S\ref{s:lst} presents the Extender in more detail, and \S\ref{s:inference} discusses inference considerations. \S\ref{s:eval} compares Extender accuracy vs. a Reference Transformer on short-context (DCLM CORE) and long-context (RULER) tasks, and \S\ref{s:conclusion} concludes. Additional results and analysis, including GQA, knowledge distillation and scaling results are available in the Appendix.

\section{Background}
\label{s:related}

Because the attention memory footprint of a standard softmax multi-head attention Transformer grows linearly with model width $d_{model}$, layer depth $L$ and context length $T$, a variety of memory-reduction approaches have been proposed.
The Extender reduces {\it persistent} attention memory: the memory required between turns, while remaining compatible with existing approaches during a turn.
That said, because a method that applies during a turn applies equally between turns, we briefly review the related work below.

\paragraph{Dimensionality Reduction}
With grouped query attention (GQA) \cite{ainslie2023gqa}, an attention layer with $H$ query heads may maintain only $H/4$ or $H/8$ keys and values per token. This in turn reduces attention memory requirements by 4--8$\times$, at a modest accuracy loss vs. full MHA.
Multi-head Latent Attention (MLA) \cite{mla} instead stores a compressed version of the $\mathbf{kv}$ pair, as much as 10$\times$ smaller.
This is decompressed on-demand before attention. As a result, MLA does not need to reduce the number of $\mathbf{kv}$ heads, but may suffer compression losses instead.

The Extender instead changes what the attention layer reads. However, GQA can be applied to the Extender to reduce memory footprint during turns (see Appendix \ref{s:gqa}).

\paragraph{Layer Reuse}
YOCO and LCKV move toward one shared cache \cite{sun2024yoco,wu2024lckv}: YOCO reuses a single global KV cache in its cross-decoder. LCKV has most layers use the same first layer KV cache.  Cross-Layer Attention (CLA) and MLKV instead share one cache among a group of layers \cite{brandon2024cla,zuhri2024mlkv}.
MiniCache instead merges similar layers' KV states after training \cite{minicache}.

The Extender significantly reuses attention input data, but does not change the attention mechanism. If desired, CLA may be applied to the Extender to save attention memory during turns. We do not evaluate this here.

\paragraph{Quantization}
With quantization \cite{xiao2023smoothquant,frantar2023gptq,liu2024kivi}, individual features or weights are represented using fewer bits, leading to both reduced memory footprint and often improved computational throughput. Applied judiciously, the accuracy loss from quantization can be modest, and savings on the order of 2--4$\times$ are not uncommon. We do not evaluate quantization in this paper.

\subsection{Approaches beyond softmax attention}
Softmax attention incurs a cost that is linear in the context length and number of layers, per token. Several alternative approaches \cite{katharopoulos2020linear,sun2023retnet,gu2023mamba,peng2023rwkv,peng2024rwkv6,goldstein2024goldfinch} have been proposed which reduce or eliminate the need to revisit each preceding token during attention.
The Extender preserves softmax attention, but changes the input to the attention $\mathbf{kv}$ projections, to achieve a reduction in memory footprint between turns.

\begin{table*}[h]
\centering
\small
\setlength{\tabcolsep}{6pt}
\begin{tabular}{lp{0.5\textwidth}}
\toprule
\textbf{Step} & \textbf{Explanation and Motivation} \\
\midrule
$\mathbf{x}_{-1}=\mathbf{h}_{-1}=[\embed(\mathrm{token})]$  &
  Both $\mathbf{x}_{-1}$ and $\mathbf{h}_{-1}$ start out as a token embedding, size $d_{model}$. $embed$ weights tied to $unembed$ below. \\

$\mathbf{s}_\ell=\Last(\mathbf{x}_{\ell-1},d_{model})$ &
  Size $d_{model}$ sliding window of $\mathbf{x}_{\ell\text{-}1}$. $\mathbf{s}_0 = \mathbf{x}_{\text{-}1}$. \\

$\mathbf{a}_\ell=\ATTN(\mathbf{s}_\ell,\mathbf{h})$ & The ATTN $\mathbf{kv}$ projections see only $\mathbf{s_\ell}$, but the $\mathbf{q}$ projection also sees $\mathbf{h}_\ell$.\\
$\mathbf{y}_\ell=\mathbf{a}_\ell+\mathbf{h}_{\ell-1}$ & $\mathbf{y_\ell}$ is the input to the FFN \\

$\hat{\delta}_\ell,\epsilon_\ell=\FFN(\mathrm{\RMSNorm}(\mathbf{y}_\ell))$ &
  The FFN outputs both the residual update $\delta_\ell$, and the extension $\epsilon_\ell$.\\

$\delta_\ell = \hat{\delta}_\ell + \mathbf{a}_\ell$ &
  $\mathbf{a}_\ell$ allows ATTN to write directly to the residual stream, and provides ATTN a clean gradient. \\

$\mathbf{x}_\ell=[\mathbf{x}_{\ell-1};\RMSNorm(\epsilon_\ell)]$ &
  Every layer appends its own read-only $\epsilon_\ell$ to $\mathbf{x}$. \\

$\mathbf{h}_\ell=\mathbf{h}_{\ell-1}+w_\ell\delta_\ell$ &
  Weighted residual stream update. The learned layer weight $w_\ell$ tends to give more weight to later layers. \\

$\mathrm{logits}=\unembed(\RMSNorm(\mathbf{h}_{L-1}))$ &
  Logits from the final hidden state. $unembed$ weights tied to $embed$ above. \\
\bottomrule
\end{tabular}

\caption{A compact description of the Extender. The extended embedding $\mathbf{x}$ corresponds to the left part of Figure \ref{f:arch}, while the residual stream $\mathbf{h}$ appears on the right. }
\label{t:archeq}
\end{table*}

\section{The Extender: A Log-Structured Transformer}
\label{s:lst}

\begin{figure*}[h]
\begin{subfigure}{0.47\textwidth}
  \centering
  \vspace{0pt}
\includegraphics[width=.8\textwidth]{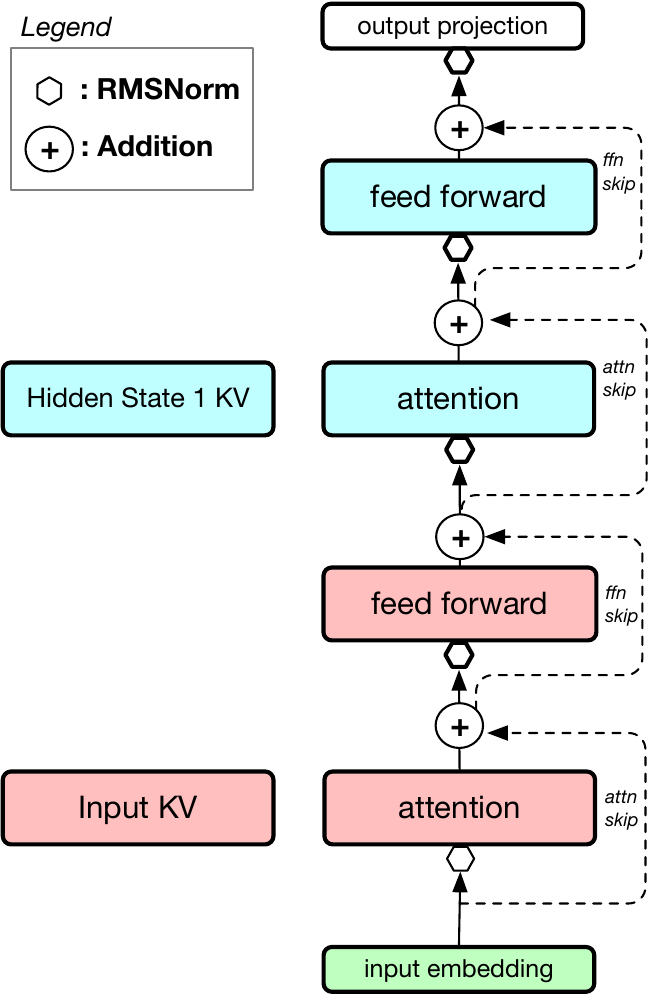}
\caption{2-layer Reference Transformer: a conventional Llama-style \cite{touvron2023llama} Transformer architecture used for comparative evaluation purposes.
Each attention layer caches unique KV-pairs derived from the input to the attention layer.}
\label{f:reftrans}
\end{subfigure}
\begin{subfigure}{0.47\textwidth}
  \centering
\includegraphics[width=0.9\textwidth]{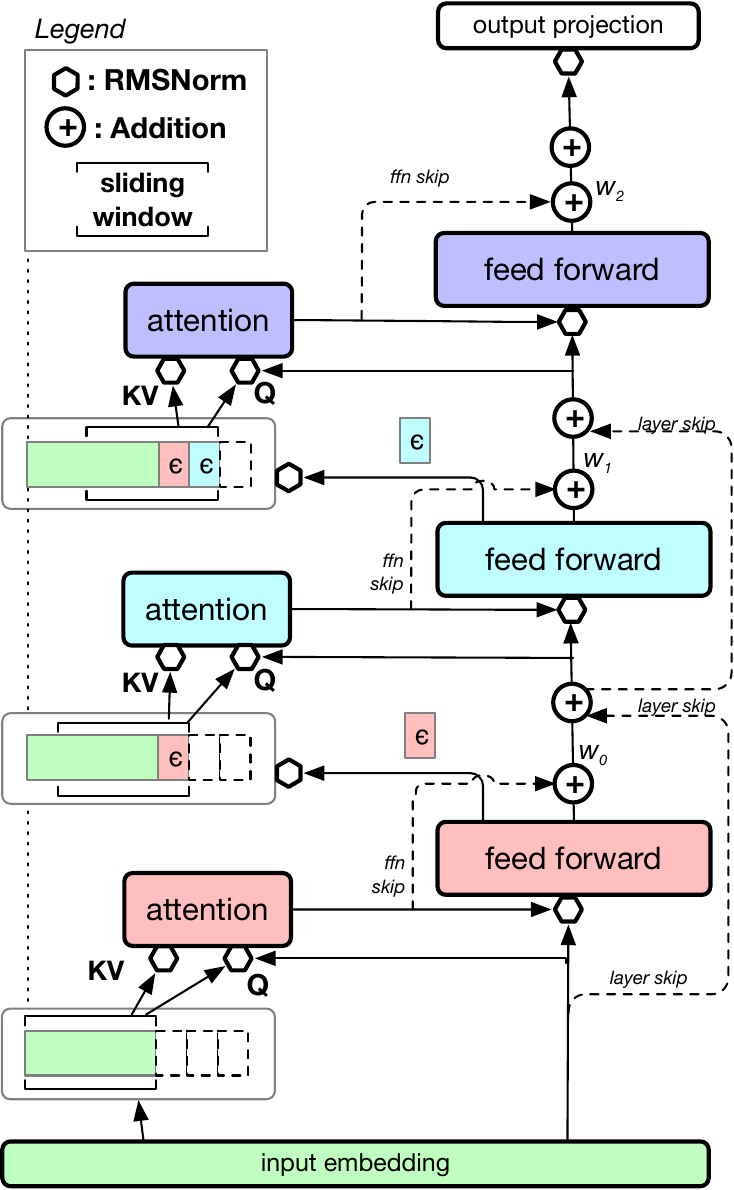}
\caption{3-Layer Extender. The extended embedding $\mathbf{x}$ grows through concatenation with the $\epsilon_\ell$ generated by each layer.
There are two layer outputs: the extension $\mathbf{\epsilon}_\ell$, and $\mathbf{\delta}_\ell$. Hexagons indicate RMSNorm.}
  \vspace{0pt}
\label{f:arch}
\end{subfigure}

\caption{Diagrams describing a conventional Transformer (left),  and an Extender (right), using the same visual language and annotations. The Extender attention $\mathbf{kv}$ projections see the log-structured extended embedding $\mathbf{x}_\ell$, rather than the residual stream $\mathbf{h}_\ell$.}
\label{f:transformers}
\end{figure*}

Figure \ref{f:arch} shows a high-level view of the Extender architecture. Compared to the Transformer in \ref{f:reftrans}, the biggest difference lies in the separation of the extended embedding $\mathbf{x}$ (left) seen by the attention $\mathbf{kv}$ projections, and the residual stream ${\mathbf{h}}$ (right) which is seen only by the FFN sub-layers and the attention $\mathbf{q}$ projections.
Key to the Extender are the {\it extensions} $\epsilon_\ell$ and {\it residual updates} $\delta_\ell$ produced by each layer.
The extended embedding $\mathbf{x}_\ell$ is {\it log-structured}, updated by concatenating $\epsilon_\ell$ to the end, rather than the addition used to update $\mathbf{h}$ by $\delta_\ell$.
After $\mathbf{x}$ and $\mathbf{h}$ have passed every layer, logits are produced from $\mathbf{h}$ only. Thus the final layer produces no $\epsilon_\ell$.

Table \ref{t:archeq} describes the model more formally, for $\ell \in 0..L\text{-}1$. Here, $unembed$ and $embed$ have tied weights.
$\mathbf{s}_\ell$ is a size $d_{model}$ window over $\mathbf{x}_{\ell-1}$. Shown here is a sliding window using a $Last()$ operator, which takes a slice of the most recently added $d_{model}$ features out of the extended embedding. In \S\ref{s:choosingepsilon} we explore alternatives to this design choice.
Note that while every $\epsilon$ persists between turns, the $d_{model}$-wide initial embedding is fixed and does not have a persistent memory footprint beyond the existing embed table and a token identifier.

Beyond the extended embedding $\mathbf{x}$, a few aspects of the Extender design bear explicit mention.
\begin{description}
\item[Dual FFN outputs] the FFN has two outputs: $\hat{\delta_\ell}$ and $\epsilon_\ell$, thus the output of the FFN is $d_{model}+d_\epsilon$ wide. The hidden dimension of the FFN is governed by the width of its input.
\item[Attention Query] the Attention query projection takes $RMSNorm(\mathbf{s}_\ell)+RMSNorm(\mathbf{h}_{\ell-1})$ as input. This does not affect the attention memory footprint, since $\mathbf{q}$ is not retained for future tokens to attend to.
\item[FFN skip] The attention output is added after a linear weight to $\mathbf{h}$. This mirrors the design of a Llama-style transformer, and allows the Attention layer to learn directly from the loss.
\end{description}

\subsection{Choosing $d_\epsilon$ and Designing the $\mathbf{kv}$ Windows}
\label{s:choosingepsilon}

In this paper, we use $|\mathbf{s}|{=}d_{model}$. This makes the Extender attention layer shape identical to that of the Transformer, making direct comparison easier.
Given this constraint, we must choose $d_\epsilon$, or more generally $|\epsilon_\ell|$.
Here, we decompose that into setting $d_\epsilon$ and $\ell_{max}$, where $\ell_{max}$ is the final layer that emits an $\epsilon$. By default, $\ell_{max}{=}L{-}2$.
Moreover, since for $\ell > 0$,  $|\mathbf{x}_\ell| \ge d_{model}$ each layer $\ell > 0$ can only read a subset of $\mathbf{x_\ell}$.

These two design decisions are closely tied: with a small $d_\epsilon$, it becomes more important that all the features of $\epsilon$ are seen by subsequent layers.
Similarly, with a large $d_\epsilon$, late layer writes may crowd out the initial embedding and early layer writes, making late layers unable to see critical features.
For example our 26-layer model has $d_{model}{=}1664$ ($\frac{\mathrm{width}}{\mathrm{height}}{=}64$). With $d_{\epsilon}{=}32$, $\sum |\epsilon_\ell|{=}800$ features, leaving room for 864 input embedding features in $\mathbf{s}_{25}$. with $d_\epsilon{=}64$, $\sum |\epsilon_\ell|{=}1600$ features. Empirically, sizes larger than 64 and smaller than 32 perform significantly worse, and are not evaluated here.

With $d_\epsilon{=}64$, the choice of read window is particularly important. Here, a sliding window leaves the final layer with only 64 features of the original embedding.
Three choices we considered were ${First}$, ${Last}$, and ${Random}$, for both the $\mathbf{k}$ and the $\mathbf{v}$ projection.
For $\mathbf{k}$, we found that ${Last}$ is always the better choice.
For $\mathbf{v}$ however, the choice is less obvious: many important tasks involve an identity mapping.
Thus $\mathbf{v}$ must be able to recall enough of the input embedding to produce the correct logit. For these, ${Last}$ would be a poor choice when $d_\epsilon{=}64$
For many other tasks, ${First}$ would instead be a poor choice.
For $d_\epsilon{=}64$ and $\ell_{max}{=}L{-}2$, $\mathbf{k}{=}Last$, $\mathbf{v}{=}Random$ may be the better compromise.
Another possibility is changing $\ell_{max}$, preventing later layers from writing to $\mathbf{x}$. For example, setting $\ell_{max}{=}L/2$ cuts $\sum |\epsilon_\ell|$ by half, leaving room for the embedding. We evaluate a few of these choices in Appendix \ref{s:choosingepsiloneval}. Our default model uses $d_\epsilon{=}32$, $\ell_{max}{=}L-2$, except
$|\epsilon_0|{=}64$.

\section{Inference and the \xstar-Cache}
\label{s:inference}

By replacing the \kvcache with an \xstarcache and computing $\mathbf{kv}$ from $\mathbf{x_\ast}$ during attention, the Extender could achieve the memory savings claimed above both during and between turns.
However, this would incur the compute cost of repeatedly reprojecting $\mathbf{kv}$ from \xstar.
To address this, the Extender uses an ephemeral \kvcache{}, which may be freed between turns as desired. This effectively amortizes the cost of reprojecting the \kvcache from the \xstarcache.
It is also worth noting that because the reprojection cost grows linearly with the total $\mathbf{kv}$ dimension, using GQA for attention dimensionality reduction \kvcache{} speeds up the reprojection time by a factor similar to the GQA space savings.

\section{Evaluation}
\label{s:eval}
\label{s:reftrans}

Below, we present results from Extender and Transformer language models trained on the ClimbMix \cite{climbmix} and ProLong \cite{prolong} datasets, using the short-context DCLM CORE \cite{core} and long-context RULER \cite{ruler} benchmark suites.

\paragraph{Reference Transformer}
In order to better understand the strengths and weaknesses of the Extender, we compare it to a {\it Reference} Transformer.
The Reference Transformer is a Llama-style MHA transformer with two small changes, following Andrej Karpathy's nanochat \cite{nanochat} design: the Muon optimizer \cite{muonoptimizer} instead of AdamW for the large matrices, and SoftCap \cite{gemma2} in the attention layer (included with Flash Attention-2 \cite{flash2}).

The Reference Transformer is a decoder-only Llama-style Transformer with a standard pre-norm residual stream.
Token embeddings of dimension $d_{model}$ are processed by \(L\) identical blocks, each of the form
\begin{align}
 \mathbf{h}\leftarrow \mathbf{h} + \mathrm{Attn}(\mathrm{RMSNorm}(\mathbf{h})),~~~~~ \mathbf{h} \leftarrow \mathbf{h} + \mathrm{FFN}(\mathrm{RMSNorm}(\mathbf{h})),
\end{align}
  followed by a final \(\mathrm{RMSNorm}\) before unembedding.
\(\mathrm{RMSNorm}\) uses \(\varepsilon {=} 10^{-5}\) and a learnable scale.

The Extender mirrors the Reference Transformer as closely as possible in terms of shape, activation functions, optimizer, training schedule, weights, norms etc.
Below are several common characteristics.

By default, we use common multi-head causal attention, with
Rotary Position Embedding (RoPE, default \(\theta {=} 10^{6}\)) applied to queries
and keys, using the standard half-dimension split.
GQA is addressed briefly in Appendix \ref{s:gqa}
Attention logits are soft-capped as \(s \cdot \tanh(z/s)\) with default
\(s = 50\), using a FlashAttention-2 kernel.
The feed-forward network is SwiGLU \cite{shazeer2020glu}, $$\mathrm{FFN}(h) = W_2\bigl(\mathrm{SiLU}(W_1\mathbf{h}) \odot W_3\mathbf{h}\bigr),$$
with hidden width \(\lfloor 8d/3\rfloor\) rounded up to a multiple of \(32\), and all linear layers bias-free.

The default tokenizer is nanochat's byte-level BPE, vocabulary 32k. Documents are prefixed with a beginning-of-sequence token, and we use document masking for
best-fit packed training sequences.

Input embeddings and the output projection share a single weight matrix
(tied embeddings).
Parameters are initialized as: embedding rows \(\mathcal{N}(0, 1/\sqrt{d})\);
multi-dimensional weights \(\mathrm{Uniform}(-1/\sqrt{\mathrm{fan\_in}},
1/\sqrt{\mathrm{fan\_in}})\); \(\mathrm{RMSNorm}\) scales are initialized to 1.

Models are trained using a WSD training schedule \cite{haegele2024wsd}.
We use the Muon optimizer \cite{muonoptimizer} for the large weight matrices, and the AdamW optimizer \cite{adamw} for the embedding weights and other one-dimensional tensors, such as norms.
Effective batch for ClimbMix 2k was 256 sequences, scaled peak learning rates were: Muon $2.1\times10^{-2}$, AdamW $1.41\times10^{-2}$ on embeddings, and $8.5\times10^{-4}$ on other 1-D parameters. For ProLong 64k, effective batch was 4 sequences, scaled peak LR: Muon $8.8\times10^{-4}$ and AdamW $3.5\times10^{-5}$ on both AdamW groups.
For efficiency, GEMMs run in bfloat16; RoPE \cite{su2024roformer} frequencies and RMSNorm \cite{zhang2019rmsnorm} statistics stay in fp32.

The design intentionally omits QK-norm, value embeddings, attention residuals, and output logit softcap, all of which could be adapted to the Extender, but would not help characterize the difference between the two architectures. The model parameter counts vary slightly between our Extender and Reference Transformer models due to architecture differences, but model width $d_{model}$ and depth $L$ are identical, and as follows. We've made an effort to use the actual sizes in the tables, plots and analysis, but will use the size class name when discussing multiple models of the same size class but different exact sizes.
\begin{table}[h]
  \centering
\begin{tabular}{rccccc}
  \bf Size & \bf Width & \bf Depth & \bf Q/K/V & \bf Head  \\
  \bf class & \bf $d_{model}$ & \bf $L$ & \bf heads & \bf dim \\
  \hline
  198M & 1024 & 13 & 8 & 128\\
  436M & 1280 & 20 & 10 & 128\\
  920M & 1664 & 26 & 13 & 128\\
\end{tabular}
\caption{Model shape for each size class. The 436M and 920M models have width-to-depth ratio of 64. The 198M model departs slightly from this ratio, to retain 128-feature heads.
The size classes are named after the Transformer size. The Extender versions have approximately 0.5\% more parameters.
}
\end{table}

We apply an auxiliary regularization cost to each token to prevent runaway residual writes: $\frac{1}{L}\sum_\ell{\mathrm{relu}(\mathrm{RMS}(\delta_\ell)-\tau)^2}$ with $\tau{=}8$ and coefficient $3{\times}10^{-3}$. After this, later layers still receive more weight ($w_\ell$ is often $10\times$ that of the earliest layers), but features of $\delta_\ell$ are in a healthy range.

\subsection{Short Context: DCLM CORE Benchmark suite}
\label{s:core}

\begin{figure*}
  \centering
  \includegraphics[width=.9\textwidth]{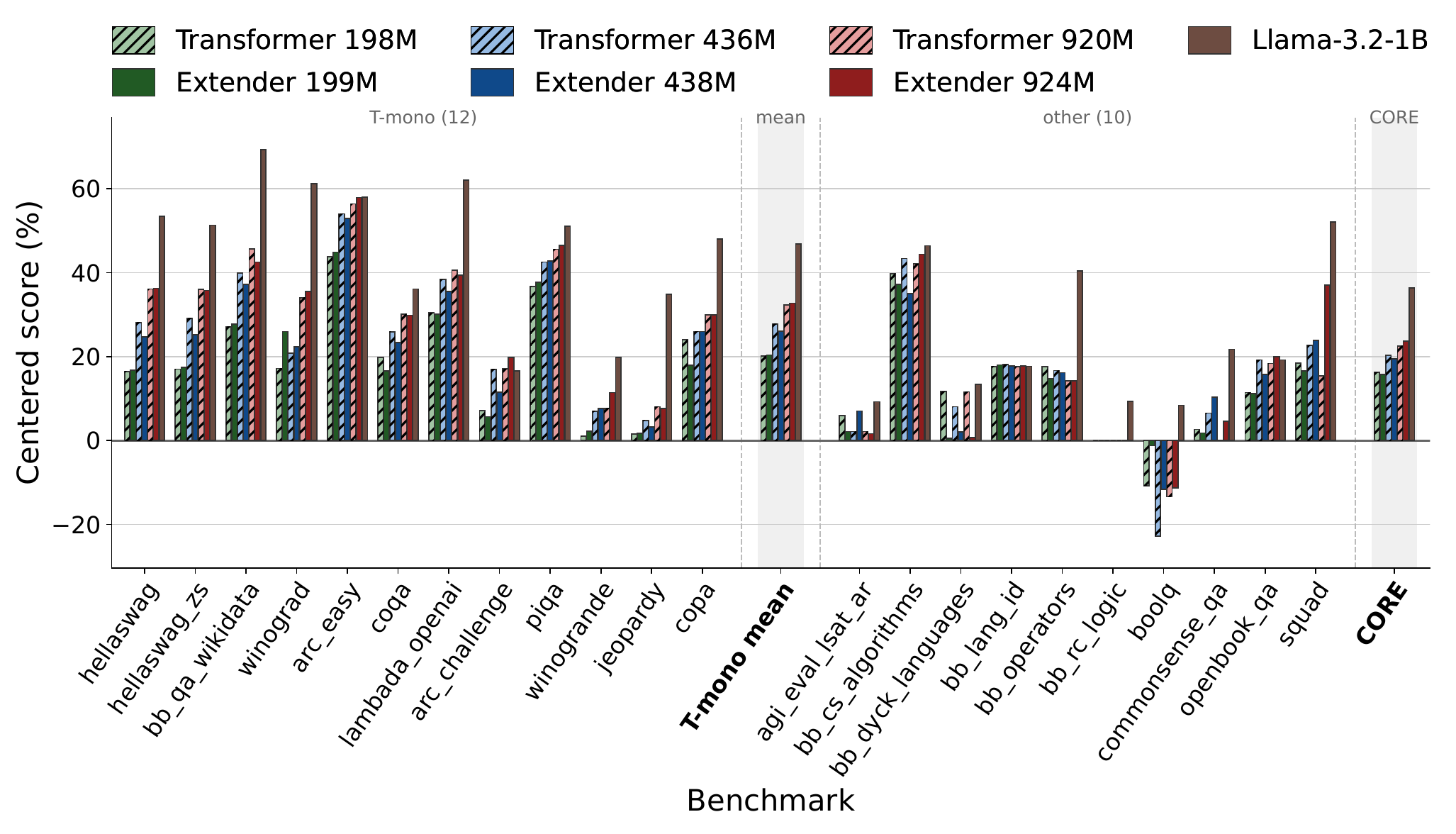}
  \caption{Centered DCLM Core v1 scores. One Transformer (hatched) and one Extender per size class, plus Llama-3.2-1B. 198M/436M/920M trained to Chinchilla token budgets. T-Mono: tasks where the Transformer score increased monotonically with parameter count under Chinchilla.}
  \label{f:core}
\end{figure*}

We trained one Extender and one Reference Transformer of each size class, from scratch, using Chinchilla \cite{chinchilla} token budgets of $\approx$20 tokens per model parameter, or approximately 4B, 9B and 18B tokens for the 198M, 436M and 920M parameter models respectively.
No fine-tuning or other task-specific training was applied.
Figure \ref{f:core} shows the performance of these models on the DCLM CORE v1 \cite{core} benchmark suite, using the nanochat \cite{nanochat} CORE harness.
We group the 22 CORE benchmarks into categories {\it T-Mono}/{\it other}. Here, {\it T-Mono} is the subset of tasks on which the Transformer score improved monotonically with parameter count under Chinchilla-matched training.
Intuitively, monotonic improvement with size suggests a more reliable benchmark in this size range.
For reference, we also include Llama-3.2-1B scores. This model was pretrained on up to 9 trillion tokens \cite{llama32}, explaining the large performance gap on some tasks. On other tasks, such as arc\_easy, Llama's 500$\times$ more training does not improve performance, suggesting the 1B model size is the primary limitation rather than the training or the architecture.

Overall, we find that the Extender and Transformer models perform similarly on DCLM CORE tasks, given parameter counts within 1\% of each other, and equal training budgets.

\begin{table*}[t]
  \centering
  \footnotesize
  \setlength{\tabcolsep}{2.36pt}
  \renewcommand{\arraystretch}{0.82}
  \caption{Mean accuracy (\%) after each mid-train stage, for Extender and Transformer (920M unless noted).
  SCORE is NIAH-6 (six needle tasks: S1--S3, MK-1, MV, MQ), REST (the other seven RULER tasks: MK-2/3, VT, CWE, FWE, QA-1/2), or RULER (13-task mean).}
  \label{tab:ruler_stages}
  \begin{tabular}{lllrrrrrrrcc}
    \toprule
    Model & Stage & SCORE & 1k & 2k & 4k & 8k & 16k & 32k & 64k & T-mono & CORE \\
    \midrule
    Extender & ClimbMix 18B & NIAH-6 & 91.4 & 82.8 & \textit{57.6} & \textit{37.2} & \textit{13.3} & \textit{4.0} & \textit{0.7} & 32.7 & 23.7 \\
             &                       & REST   & 31.2 & 22.5 & \textit{11.6} & \textit{5.0} & \textit{4.1} & \textit{3.7} & \textit{2.9} & & \\
             &                       & RULER  & 59.0 & 50.3 & \textit{32.9} & \textit{19.9} & \textit{8.4} & \textit{3.8} & \textit{1.9} & & \\
    \addlinespace
    Transformer & ClimbMix 18B & NIAH-6 & 90.0 & 84.4 & \textit{17.5} & \textit{0.0} & -- & -- & \textit{--} & 32.3 & 22.5 \\
                &                       & REST   & 33.1 & 24.1 & \textit{5.9 }& \textit{0.0} & -- & -- & \textit{--} & & \\
                &                       & RULER  & 59.4 & 51.9 & \textit{11.2} & \textit{0.0} & -- & -- & \textit{--} & & \\

                \midrule

    Extender & ProLong 64k $+2.5$B & NIAH-6 & 92.8 & 87.6 & 85.1 & 77.5 & 78.6 & 73.2 & 43.3 & 30.8 & 23.5 \\
             &                       & REST   & 51.4 & 43.1 & 31.1 & 23.7 & 19.2 & 7.9 & 4.8 & & \\
             &                       & RULER  & 70.5 & 63.7 & 56.1 & 48.5 & 46.6 & 38.0 & 22.6 & & \\
    \addlinespace
        Transformer & ProLong 64k $+2.5$B & NIAH-6 & 91.0 & 83.6 & 75.4 & 55.7 & 48.9 & 42.9 & 21.0 & 29.9 & 21.8 \\
                &                       & REST   & 38.5 & 18.6 & 11.1 & 8.2 & 7.6 & 5.0 & 3.0 & & \\
                &                       & RULER  & 62.7 & 48.6 & 40.8 & 30.1 & 26.7 & 22.5 & 11.3 & & \\
    \midrule

    Extender & prolong 64k $+5$B & NIAH-6 & 90.7 & 84.3 & 83.3 & 75.0 & 78.6 & 74.7 & 59.3 & 31.5 & 22.8 \\
             &                       & REST   & 49.9 & 40.0 & 41.4 & 28.3 & 28.6 & 17.3 & 9.9 & & \\
             &                       & RULER  & 68.7 & 60.4 & 60.8 & 49.9 & 51.7 & 43.8 & 32.7 & & \\
    Transformer & ProLong 64k +5B & NIAH-6 & 87.8 & 84.3 & 84.0 & 71.0 & 65.3 & 64.6 & 28.5 & 30.8 & 21.8 \\
                &                       & REST   & 42.2 & 29.7 & 20.4 & 14.4 & 17.2 & 9.9 & 7.1 & & \\
                &                       & RULER  & 63.2 & 54.9 & 49.8 & 40.5 & 39.4 & 35.1 & 17.0 & & \\

    \bottomrule
  \end{tabular}
\end{table*}

\subsection{Long Context: RULER}
\label{s:ruler}

\begin{figure*}
  \centering
  \includegraphics[width=\linewidth]{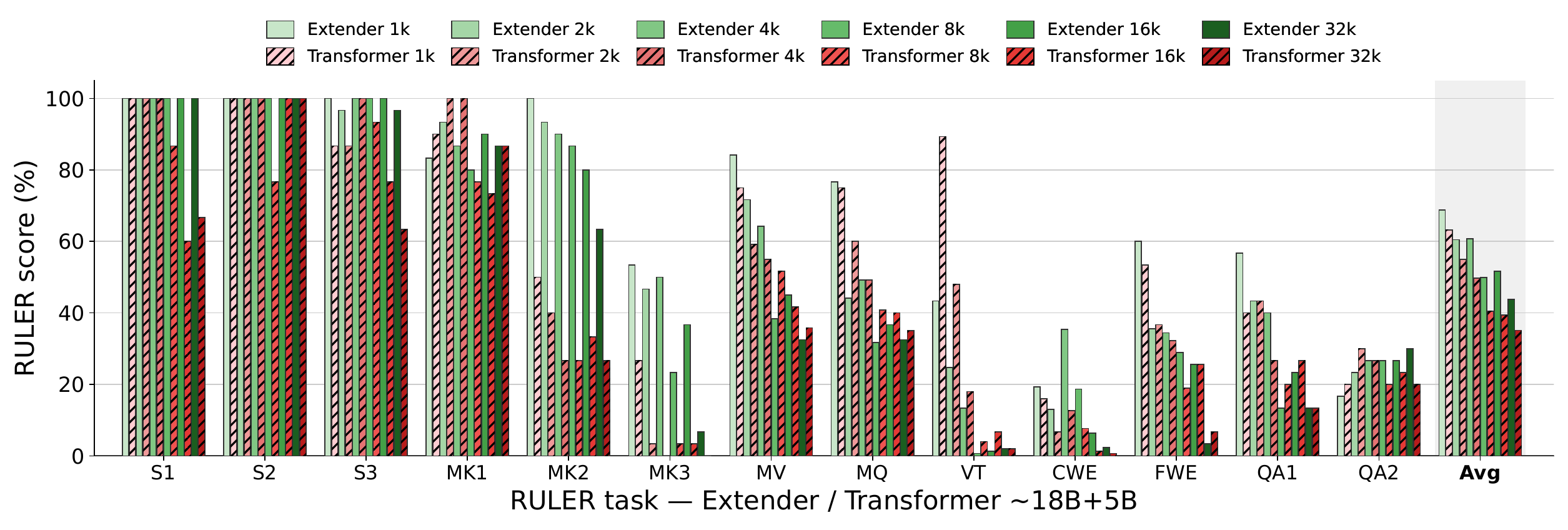}
  \caption{Per-task RULER scores for the ClimbMix 18B $\to$ $+5$B 64k ProLong checkpoints, at evaluation lengths 1k / 2k / 4k / 8k / 16k / 32k.
  Extender is green, Transformer is red; darker bars are longer contexts.
  All 13 tasks are shown. Avg is the 13-task RULER mean.}
  \label{f:ruler5b}
\end{figure*}

In this section, we evaluate Reference Transformer and Extender performance on long-context tasks using the RULER \cite{ruler} benchmark suite.
The purpose of the evaluation is an apples-to-apples comparison between Extender and Transformer architectures, not to maximize RULER scores.
Models were not fine-tuned for instruction-following or RULER. As a result, these absolute scores are lower than overtrained or fine-tuned public models.

Here, we focus on the 920M parameter models, and evaluate them at the pretrain stage, after 2.5B tokens of 64k ProLong data \cite{prolong}, and after 5B tokens of 64k ProLong respectively, using the recommended 60/40 mix of long vs. short context documents.
For 2.5B ProLong training, we kept $\theta=10^6$, and resumed training from the pre-decay pretrain checkpoint, then trained with ProLong tokens finished by a brief decay.
For 5B ProLong training, we similarly resumed training from the pre-decay checkpoint of the 2.5B training run.
The resulting aggregate token count for the 5B ProLong models is 21.3B tokens (some ClimbMix tokens lost here due to pre-decay resume).

Figure~\ref{f:ruler5b} shows how the Extender and the Reference Transformer performed on RULER for context lengths between 1k and 32k tokens. Results are based on $n{=}30$ samples per task, with greedy decoding. On average, we find that the Extender outperforms the Reference Transformer on these long-context tasks.
Two interesting details stand out: the Extender has a large advantage on
the harder multi-key tasks, while the Transformer wins convincingly at the variable tracking task. This suggests a possible difference in inductive bias on these three tasks. However, we did not investigate this further.

We define a subset NIAH-6 as the six easier tasks: single-key needle-in-a-haystack variants 1--3 (S1--S3), multivalue (MV), multiquery (MQ), and multikey variant 1 (MK1).
We use the label REST to refer to the remaining group of tasks (MK2--3, VT, CWE, FWE, QA1 and QA2).
Table~\ref{tab:ruler_stages} reports NIAH-6, REST, and the 13-task RULER mean at each evaluation length, plus T-mono and centered CORE (\%) for the ClimbMix 2k base and the 64k ProLong mid-train stages.
After the ClimbMix pretrain, and before ProLong training, the Extender is significantly better than the Transformer at context lengths beyond the 2k training context. For 1-2k contexts, any differences are marginal.
After ProLong training, the Reference Transformer improves significantly on longer contexts over its pretrain equivalent. However, Transformer performance on long-context tasks consistently trails the Extender, on both the easier NIAH-6 and the harder REST tasks.

For the 920M models shown, the persistent attention memory footprint at 64k context is approximately 54.5 million features (109 MB at bf16) for the Extender, vs. 5.67 billion features (11.3 GB at bf16) for the (MHA) Reference Transformer, a factor $104\times$ reduction.
Despite this, we find the Extender to be interchangeable with the Transformer on short context DCLM CORE tasks, and somewhat ahead on long-context RULER tasks.

\subsection{Ablation Studies}

\begin{table*}[h]
  \centering
  \small
  \setlength{\tabcolsep}{4pt}
  \caption{Extender at $d_\epsilon{=}32$
  ($\sim$200\,M, $d{=}1024$, $L{=}13$, $\approx$4\,B ClimbMix tokens, seed 42).
  Each later row changes one default choice.}
  \label{tab:ablation_eps32}
  \begin{tabular}{lccccccc}
    \toprule
    Configuration & Val. & T-mono & CORE & \multicolumn{4}{c}{RULER} \\
    \cmidrule(lr){5-8}
     & & & & 1k & 2k & 4k & 8k \\
    \midrule
    default Extender & 2.7547 & 20.1 & 14.6 & 50.1 & 39.3 & 25.7 & 13.5 \\
    \midrule
    $Q$ from $\mathrm{RMSNorm}(h)$ & 2.7555 & 20.4 & 15.7 & 47.0 & 41.8 & 12.6 & 1.2 \\
    $Q$ from $h$ & 2.7617 & 19.5 & 15.1 & 44.7 & 41.5 & 22.1 & 12.4 \\
    $Q$ from $x$ & 2.7695 & 19.2 & 15.4 & 44.9 & 38.5 & 18.1 & 8.0 \\
    $Q$ from $\mathrm{RMSNorm}([x;h])$ & 2.7500 & 20.2 & 15.6 & 48.0 & 37.0 & 25.9 & 13.9 \\
    $Q$ from $[x;h]$& 2.7586 & 20.0 & 14.3 & 48.2 & 39.4 & 23.1 & 10.6 \\
    \midrule
    FFN input: $\mathrm{RMSNorm(\mathbf{a_\ell})}$ & 2.8398 & 17.8 & 14.5 & 42.4 & 35.3 & 25.3 & 10.2 \\
    FFN input: $\mathrm{RMSNorm(\mathbf{a_\ell}+RMSNorm(\mathbf{h}))}$ & 2.7727 & 19.6 & 14.8 & 46.0 & 39.2 & 33.0 & 15.4 \\
    \bottomrule
  \end{tabular}
\end{table*}

To better understand how each aspect of the Extender contributes to its performance, we run a series of ablation studies.
For these, we train matched $\sim$200\,M Extenders ($d_{\mathrm{model}}{=}1024$, $L{=}13$, $d_\epsilon{=}32$) on ClimbMix for $\approx$4\,B tokens with a shared WSD recipe, varying one architectural choice at a time relative to the default Extender residual graph. Table~\ref{tab:ablation_eps32} reports final validation loss, DCLM CORE (T-mono mean and overall CORE, centered) and RULER, for a subset of the ablation tests performed.

\paragraph{Attention Query Input} In contrast to the Transformer, Extender has a choice of what input to take for the attention query projection matrix: $\mathbf{x}$, $\mathbf{h}$, or some combination of the two, with or without RMSNorm applied to the input. The default setting is $Q$ from $\mathrm{RMSNorm}(\mathbf{x})+\mathrm{RMSNorm}(\mathbf{h})$.
In the table, $[x;h]$ denotes concatenation rather than addition of inputs.
Many of the differences are marginal, however, two significant results stand out: $Q$ from either $h$ or $x$ alone significantly underperforms on long-context tasks.
Note that a concatenated input, with a $W_q$ that is twice as large, did not substantially improve accuracy.

\paragraph{Residual Input into FFN}
As seen in Table \ref{t:archeq}, the default configuration passes $\mathrm{RMSNorm(\mathbf{a}_\ell+\mathbf{h}_{\ell-1})}$ as input to the FFN. Here, we test two ablations: eliminating $\mathbf{h}$ as input, and applying an additional norm on $\mathbf{h}$: $\mathrm{RMSNorm(\mathbf{a}_\ell+RMSNorm(\mathbf{h_{\ell-1}}))}$.

From these ablations, it is clear that including $\mathbf{h}$ in the FFN input lifts performance. However, compared to a conventional Transformer, the loss from dropping $\mathbf{h}$ is not catastrophic: much of the communication between layers is carried by $\mathbf{x}$ rather than relying fully on $\mathbf{h}$. For out-of-domain tasks, having $\mathbf{h}$ in Q is much more important than having $\mathbf{h}$ in the FFN input. The extra RMSNorm ablation reports mixed and likely insignificant results.

\subsection{Limitations}

While the Extender with $d_\epsilon=32$ scales well from 199M to 924M parameters, we have not yet been able to test it at much larger model sizes due to resource constraints.
Similarly, our token budgets are largely limited to 20$\times$ the parameter count.

\section{Conclusion}
\label{s:conclusion}

In contrast with prior work on reducing the attention memory footprint, the Extender focuses on the persistent memory footprint: memory required between turns.

The Extender separates the superposition channel $\mathbf{h}$ focused on next-token prediction, from the concatenation channel $\mathbf{x}$ dedicated to attention memory.
For every size class we tested, 32 features of attention memory per layer was enough to match the Reference Transformer on short-context tasks, and outperform it on long-context tasks.

\clearpage

\FloatBarrier
\clearpage
\bibliographystyle{iclr2027_conference}
\bibliography{refs}
\FloatBarrier
\clearpage

\appendix
\section{Additional Long-Context RULER Evaluation Details}
\label{s:rulerdetails}

\begin{figure*}[h]
  \centering
  \includegraphics[width=\linewidth]{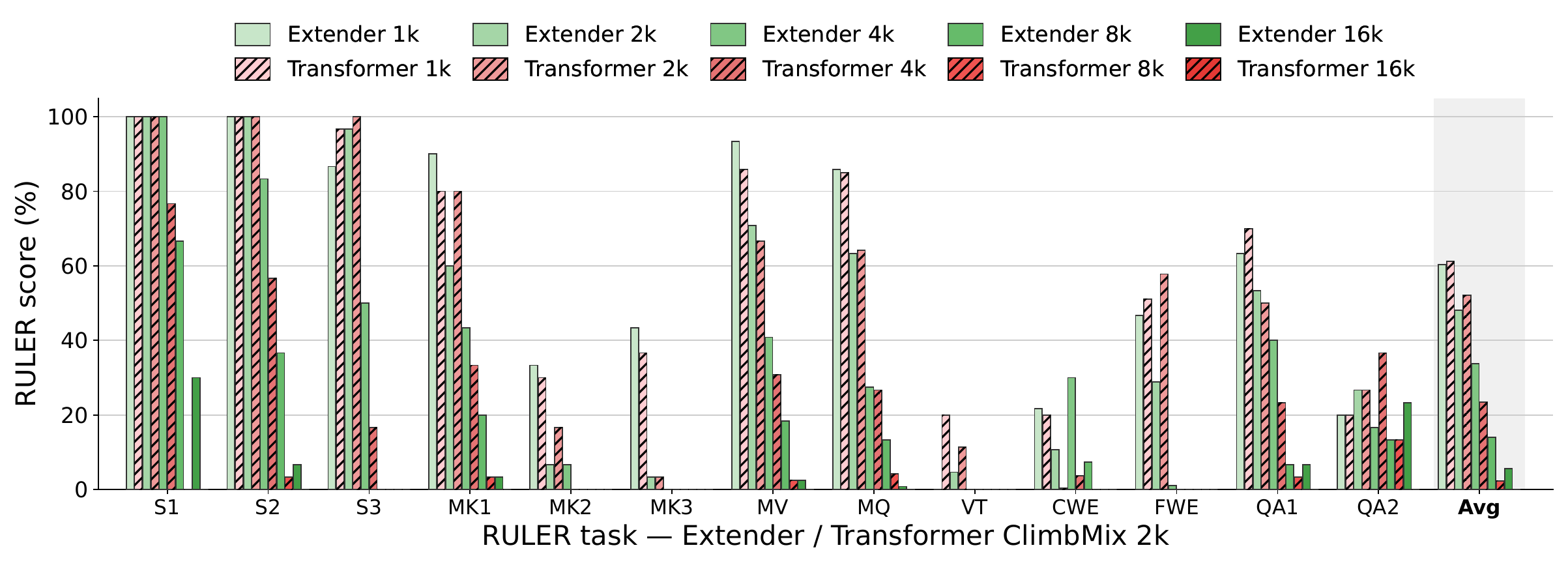}
  \caption{Per-task RULER scores for the ClimbMix 2k pretrain checkpoints, at evaluation lengths 1k / 2k / 4k / 8k / 16k.
  Extender ($d_\epsilon{=}32$) is green, Transformer is red.}
  \label{f:ruler_pretrain}
\end{figure*}

\begin{figure*}[h]
  \centering
  \includegraphics[width=\linewidth]{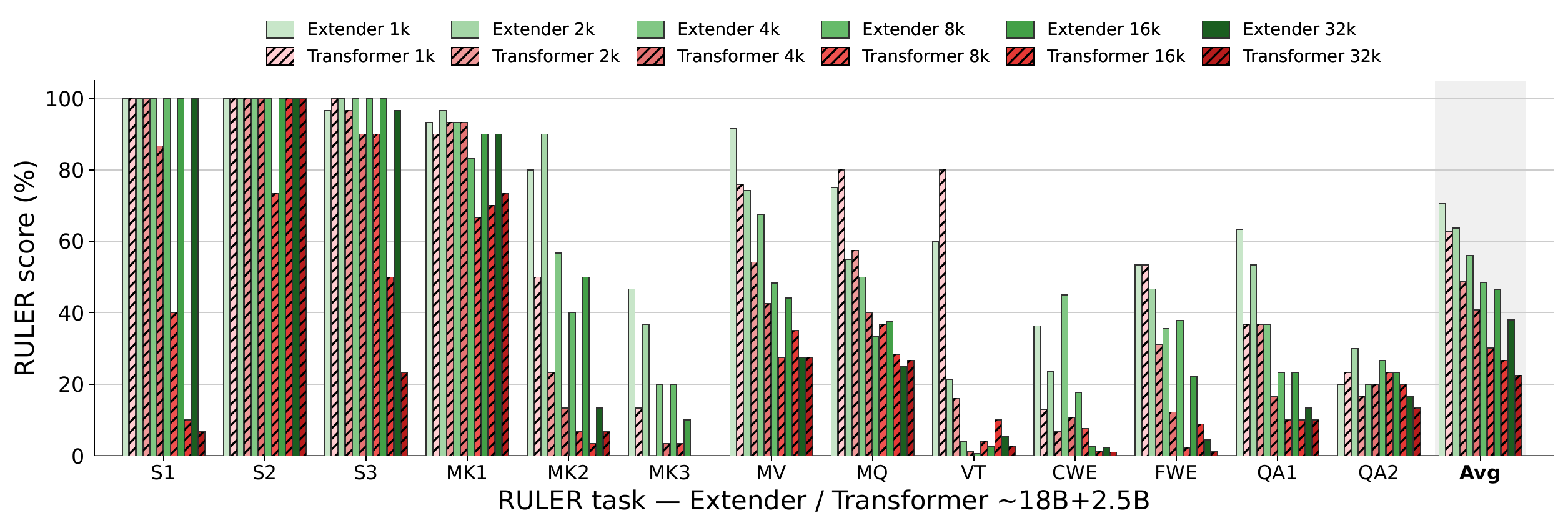}
  \caption{Per-task RULER scores for the ClimbMix 18B $\to$ $+2.5$B 64k ProLong checkpoints, at evaluation lengths 1k / 2k / 4k / 8k / 16k / 32k.
  All 13 tasks are shown. Avg is the 13-task RULER mean.}
  \label{f:ruler2p5b}
\end{figure*}

Figures \ref{f:ruler_pretrain} and \ref{f:ruler2p5b} offer additional training stage RULER performance details. In Figure \ref{f:ruler_pretrain},
the Extender has a clear advantage at context lengths beyond the 2k training context.
Figure~\ref{f:ruler2p5b} is the $+2.5$B stage of Figure~\ref{f:ruler5b}, with the Extender advantage already clearly visible.

\begin{figure*}[p]
  \centering
  \begin{subfigure}[t]{\textwidth}
    \centering
    \includegraphics[width=\linewidth]{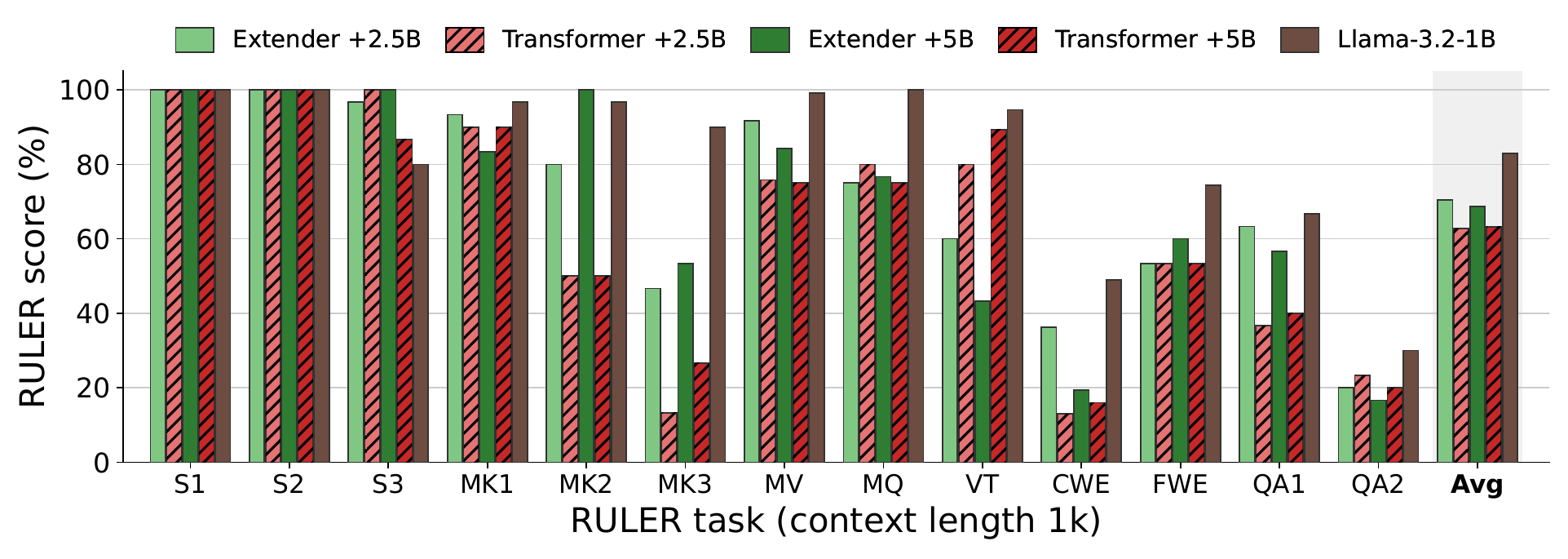}
    \caption{Context length 1k.}
    \label{f:ruler1k}
  \end{subfigure}
  \vspace{0.25em}
  \begin{subfigure}[t]{\textwidth}
    \centering
    \includegraphics[width=\linewidth]{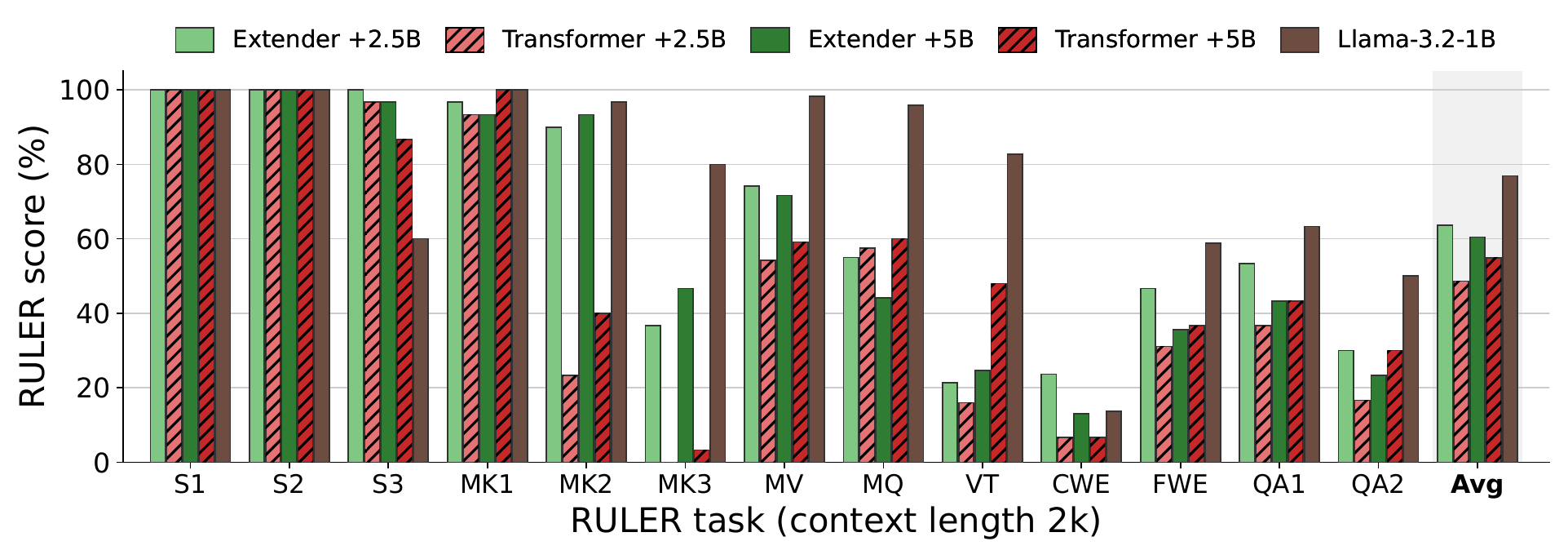}
    \caption{Context length 2k.}
    \label{f:ruler2k}
  \end{subfigure}
  \vspace{0.25em}
  \begin{subfigure}[t]{\textwidth}
    \centering
    \includegraphics[width=\linewidth]{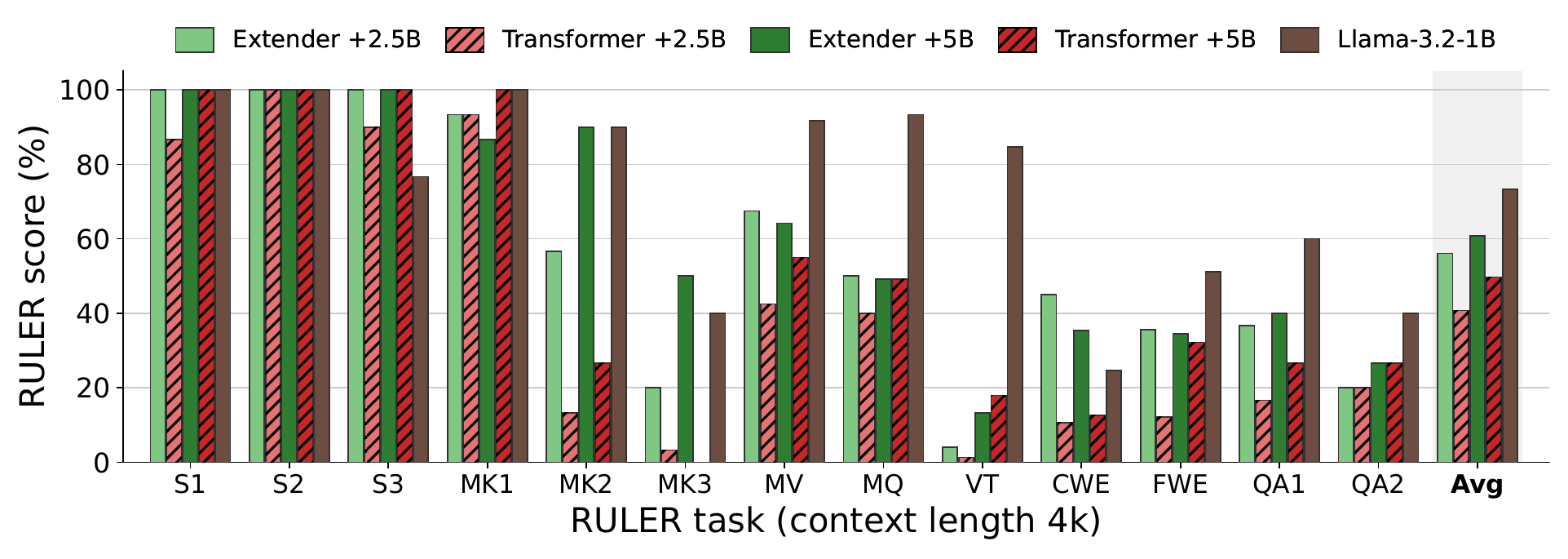}
    \caption{Context length 4k.}
    \label{f:ruler4k}
  \end{subfigure}
  \caption{Per-task RULER scores after 64k ProLong mid-train from the ClimbMix 18B checkpoint ($+2.5$B / $+5$B), at 1k, 2k, and 4k.
   Extender / Transformer: $d{=}1664$, $L{=}26$. Sizes 8k, 16k, and 32k in the next figure. }

  \label{f:rulermix1k4k}
\end{figure*}

\begin{figure*}[p]
  \centering
  \begin{subfigure}[t]{\textwidth}
    \centering
    \includegraphics[width=\linewidth]{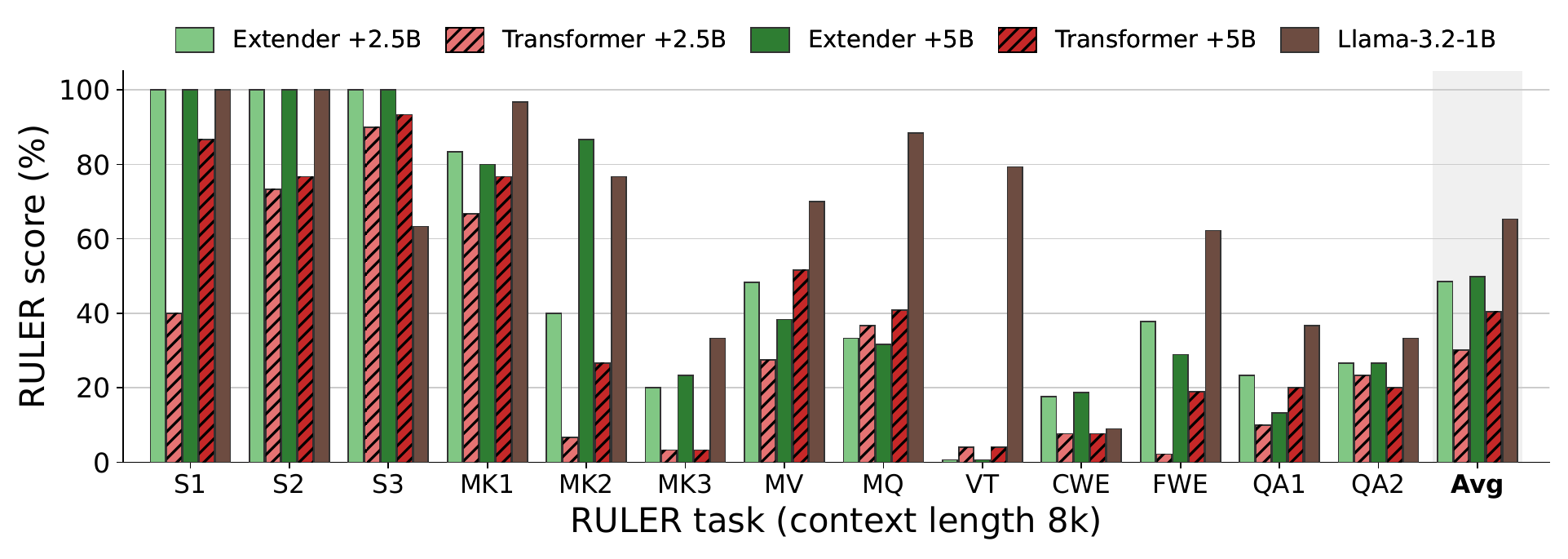}
    \caption{Context length 8k.}
    \label{f:ruler8k}
  \end{subfigure}
  \vspace{0.25em}
  \begin{subfigure}[t]{\textwidth}
    \centering
    \includegraphics[width=\linewidth]{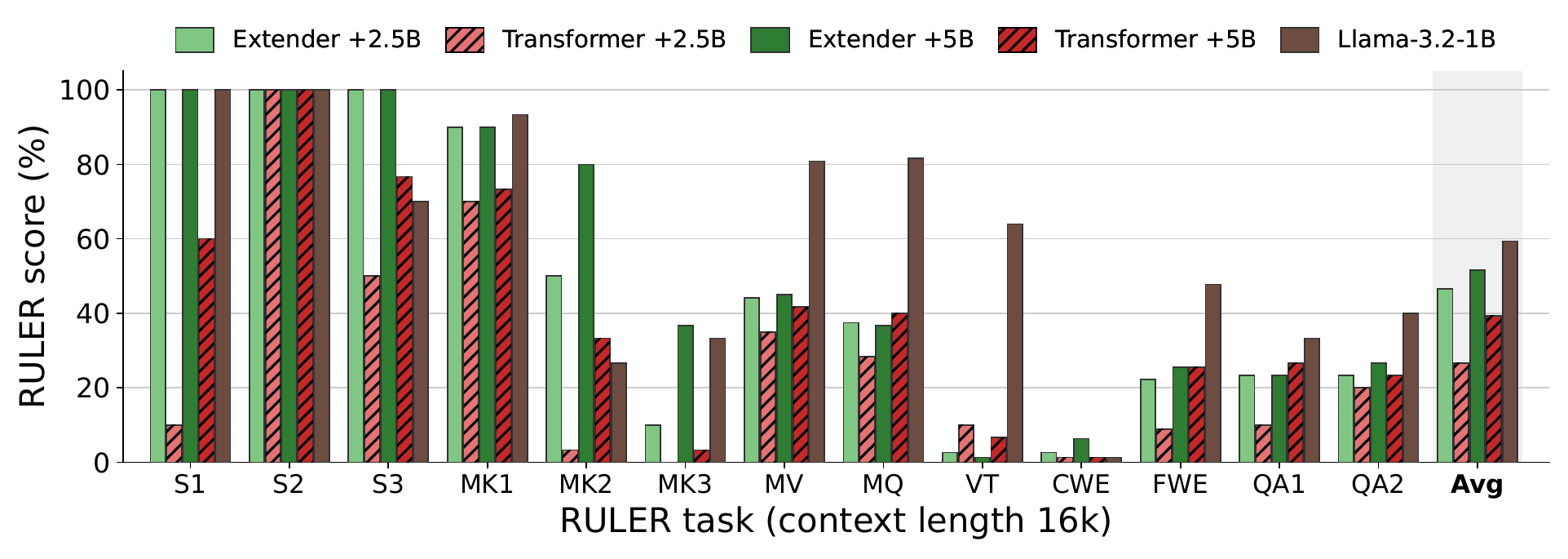}
    \caption{Context length 16k.}
    \label{f:ruler16k}
  \end{subfigure}
  \vspace{0.25em}
  \begin{subfigure}[t]{\textwidth}
    \centering
    \includegraphics[width=\linewidth]{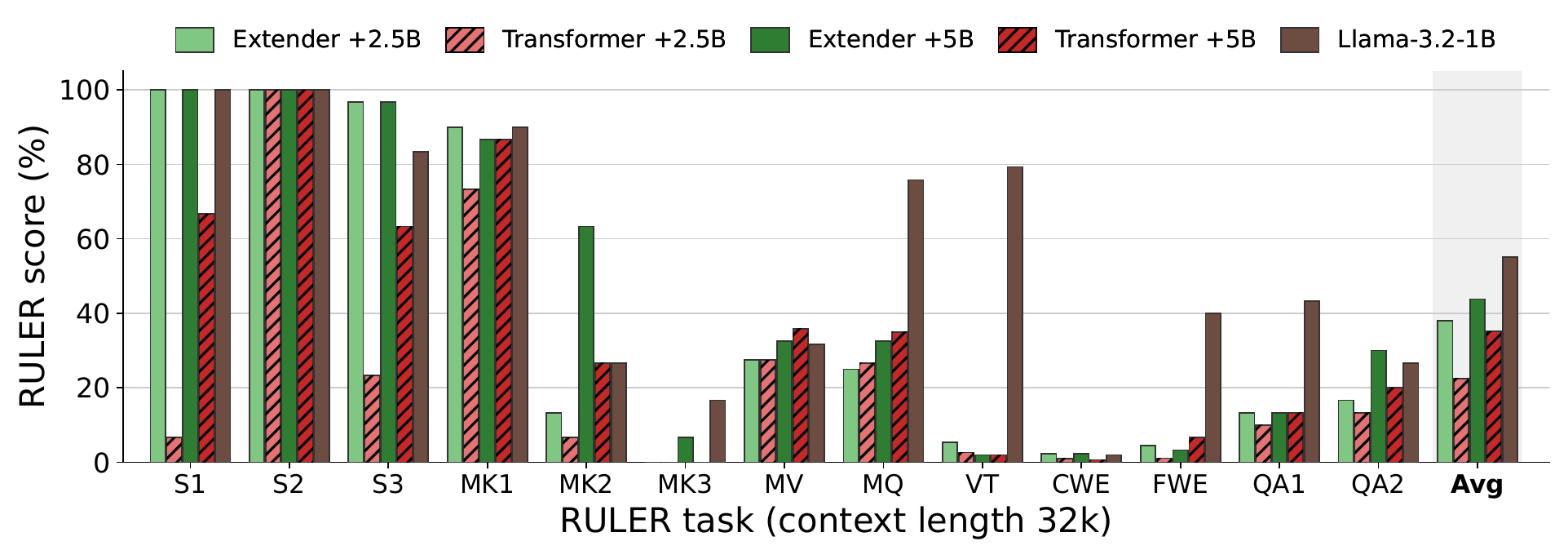}
    \caption{Context length 32k.}
    \label{f:ruler32k}
  \end{subfigure}
  \caption{Same models as Figure~\ref{f:rulermix1k4k}, at 8k, 16k, and 32k.
  }
  \label{f:ruler16k32k}
\end{figure*}

Figures \ref{f:ruler1k}--\ref{f:ruler32k} show detailed per-task RULER results per context length, up to 32k.

\FloatBarrier

\section{$\epsilon_\ell$ size and $V$ window design}
\label{s:choosingepsiloneval}

\begin{table*}[h]
  \centering
  \small
  \setlength{\tabcolsep}{4pt}
  \caption{Window and $\epsilon$ choices at 920M pretrain
  ($\sim$900\,M, $d{=}1664$, $L{=}26$, ClimbMix 2k, 9.6\,B tokens).
  wide0 sets $|\epsilon_0|{=}2 d_\epsilon$.
  $\ell_{max}=13$ uses $K$ and $V$ from $\Last(d)$ and emits $\epsilon_\ell$ on only the first 13 layers.
  RULER is the 13-task mean (\%).}
  \label{tab:lane_recipe_900m}
  \begin{tabular}{lccccccc}
    \toprule
    Configuration & Val. & T-mono & CORE & \multicolumn{4}{c}{RULER} \\
    \cmidrule(lr){5-8}
     & & & & 1k & 2k & 4k & 8k \\
    \midrule
    $d_\epsilon{=}32$ wide0, $V{=}\Last$ & 2.4983 & 30.5 & 23.0 & 60.3 & 48.1 & \textit{33.8} & \textit{14.0} \\
    $d_\epsilon{=}32$, $V{=}\Last$ & 2.4984 & 30.5 & 22.9 & 57.3 & 44.3 & \textit{36.9} & \textit{14.3} \\
    $d_\epsilon{=}64$ $\ell_{max}{=}13$ & 2.4781 & 30.0 & 22.9 & 61.9 & 47.7 & \textit{33.6} & \textit{15.1} \\
    $d_\epsilon{=}64$, $V{=}\Last$ & 2.4934 & 30.8 & 22.9 & 63.0 & 53.9 & \textit{34.6} & \textit{15.3} \\
    $d_\epsilon{=}64$, $V{=}\mathrm{Rand}$ & 2.4773 & 30.4 & 23.4 & 58.6 & 47.7 & \textit{33.8} & \textit{13.3} \\
    $d_\epsilon{=}128$, $V{=}\mathrm{Rand}$ & 2.4842 & 29.2 & 20.6 & 61.4 & 52.0 & \textit{32.8} & \textit{14.7} \\
    \midrule
    Transformer & 2.4983 & 32.3 & 23.5 & 58.5 & 46.2 & \textit{19.7} & \textit{0.8} \\
    \bottomrule
  \end{tabular}
\end{table*}

Table \ref{tab:lane_recipe_900m} evaluates $d_\epsilon$ vs. window design for $V$ on our 920M model class. Due to resource constraints, these ablation models are trained to 9B tokens instead of the full 18B. In our experience, relative 9B token model performance is a good indicator of relative 18B model performance.
We evaluate six different choices. Here, {\it wide0} doubles the width of the first layer: by iteratively zeroing out $\epsilon_\ell$ during evaluations, we have anecdotally found that $\epsilon_0$ is critical to  Extender performance on most tasks, while others are task dependent.
$d_\epsilon{=}32$ with {\it wide0} is the default Extender.
However, we find the differences between most models are marginal in these ablations, $d_\epsilon=128$ being a notable exception for short contexts.
Anecdotally, $d_\epsilon=64$ V=Rand is a strong choice when fully trained.
We did not choose it as the default model due to the cleaner and more memory and compute-efficient design of the $d_\epsilon{=}32, V{=}Last$ options.

\FloatBarrier

\section{Distillation and Extenders}
\label{s:distillation}

\begin{table}[h]
  \centering
  \small
  \setlength{\tabcolsep}{4pt}
  \caption{Logit distillation into a 198M Extender.
   T-mono and CORE are centered DCLM CORE (\%); T-mono is the 12-task set in Figure~\ref{f:core}.
   RULER is the 13-task mean (\%) at 1k--8k. }
  \label{tab:distill_200m}
  \begin{tabular}{lccccccc}
    \toprule
    Model & Tokens & T-mono & CORE & \multicolumn{4}{c}{RULER} \\
    \cmidrule(lr){5-8}
     & & & & 1k & 2k & 4k & 8k \\
    \midrule
    199M Extender, from scratch & 4B & 19.9 & 15.9 & 46.8 & 38.5 & \textit{24.4} & \textit{12.0} \\
    199M Extender student, Extender\ teacher & 4B & 24.1 & 19.1 & 51.7 & 44.0 & \textit{29.6} & \textit{14.8} \\
    199M Extender student, Transformer\ teacher & 4B & 23.8 & 17.2 & 51.9 & 44.0 & \textit{27.8} & \textit{16.0} \\
    \midrule
    924M Extender (teacher) & 14.5B & 33.2 & 25.1 & 70.7 & 64.1 & 62.5 & 53.5 \\
    920M Transformer (teacher) & 14.5B & 32.1 & 23.9 & 68.3 & 58.1 & 54.7 & 50.1 \\
    \bottomrule
  \end{tabular}
\end{table}

Knowledge distillation \cite{hinton2015distill,sanh2019distilbert,gu2024minillm} trains a student using a larger teacher model, often achieving higher token efficiency than training from scratch.
To better understand differences in inductive bias \cite{goyal2022inductive,battaglia2018relational} between Transformer and Extender models, we train three 198M class extenders: one from scratch, one with a 924M Extender teacher, and one with a 920M Transformer teacher.
Here, all three 198M Extenders use $d_\epsilon{=}64$ and $\mathbf{v}$ selects features from $\mathbf{x}$ using the $Random$ policy. All three are trained to 4 Billion ClimbMix tokens with a 2k token sequence length. Both teachers were trained with 9B ClimbMix + 5B ProLong tokens - earlier models than the 18B reported in the main body of the paper. The Extender teacher is somewhat stronger than the Transformer teacher, on both CORE and RULER scores.

Table~\ref{tab:distill_200m} shows the results of this experiment. The two student models both consistently outperform the from-scratch model, showing that Extenders train well under a distillation objective. The student led by an Extender teacher achieved a CORE score 6 points lower than its teacher, while the Transformer teacher transferred slightly less knowledge, with the student scoring 6.7 points below the teacher. On long-context RULER tasks, the two students achieved near-identical scores within the 2k training domain despite the Transformer teacher being weaker. These results indicate that under logit distillation, an Extender student learns approximately as well from an Extender teacher as from a Transformer teacher.

\FloatBarrier

\section{Grouped Query Attention and Extenders}
\label{s:gqa}

\begin{table*}[h]
  \centering
  \small
  \setlength{\tabcolsep}{3.6pt}
  \caption{GQA vs. MHA Extender and MHA Transformer
  RULER is the 13-task mean (\%).
  For all models $d_{model}=1664$. Head dimension 128 $\to$ 13 heads.
  Thus GQA4/5 is 13 query / 3 KV heads (groups $5,4,4$).
  Due to the dimensionality reduction of GQA, the parameter count of the GQA4/5 model is 813M rather than 924M.
   }
  \label{tab:gqa}
  \begin{tabular}{lccccccccc}
    \toprule
    Model & T-mono & CORE & \multicolumn{7}{c}{RULER} \\
    \cmidrule(lr){4-10}
     & & & 1k & 2k & 4k & 8k & 16k & 32k & 64k \\
    \midrule
    Extender 924M (MHA) & 32.7 & 23.7 & 59.0 & 50.3 & \textit{32.9} & \textit{19.9} & \textit{8.4} & \textit{3.8} & \textit{1.9} \\
    Extender 813M (GQA4/5) & 33.1 & 22.9 & 61.2 & 52.4 & \textit{36.3} & \textit{18.4} & \textit{6.6} & \textit{2.1} & \textit{1.2} \\
    Transformer 920M (MHA) & 32.3 & 22.5 & 59.4 & 51.9 & \textit{11.2} & \textit{0.0} & -- & -- & \textit{--} \\
    \midrule
    Extender 924M $+2.5$B (MHA) & 30.8 & 23.5 & 70.5 & 63.7 & 56.1 & 48.5 & 46.6 & 38.0 & 22.6 \\
    Extender 813M $+2.5$B (GQA4/5) & 31.2 & 23.3 & 63.9 & 57.9 & 56.3 & 49.7 & 45.8 & 33.8 & 20.2 \\
    Transformer 920M $+2.5$B (MHA)  & 29.9 & 21.8 & 62.7 & 48.6 & 40.8 & 30.1 & 26.7 & 22.5 & 11.3 \\
    \midrule
    Extender 924M $+5$B (MHA) & 31.5 & 22.8 & 68.7 & 60.4 & 60.8 & 49.9 & 51.7 & 43.8 & 32.7 \\
    Extender 813M $+5$B (GQA4/5) & 31.6 & 22.5 & 67.3 & 61.2 & 58.4 & 45.2 & 40.2 & 37.7 & 26.3 \\
    Transformer 920M $+5$B  (MHA) & 30.8 & 21.8 & 63.2 & 54.9 & 49.8 & 40.5 & 39.4 & 35.1 & 17.0 \\
    \bottomrule
  \end{tabular}

\end{table*}

\begin{figure*}[h]
  \centering
  \includegraphics[width=\linewidth]{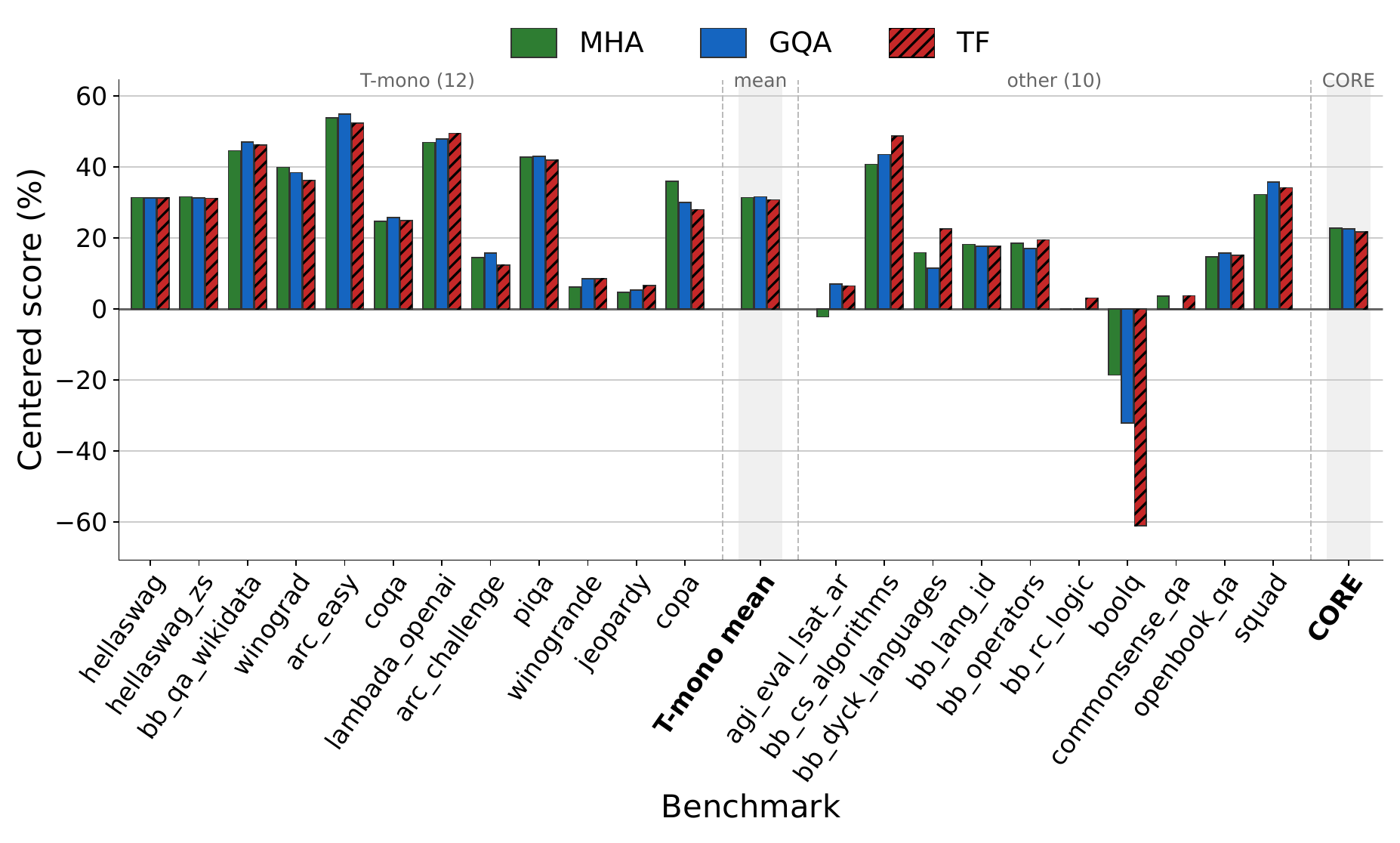}
  \caption{Centered DCLM CORE subscores for the 18B ClimbMix $+5$B ProLong checkpoints in Table~\ref{tab:gqa}.
  MHA is the full-attention Extender, GQA is 13 query / 3 KV heads (groups $5,4,4$), and TF is the document-mask Transformer (hatched).
  T-mono is the 12-task set from Figure~\ref{f:core}; CORE is the 22-task mean.}
  \label{f:gqa_core}
\end{figure*}

\begin{figure*}[h]
  \centering
  \includegraphics[width=\linewidth]{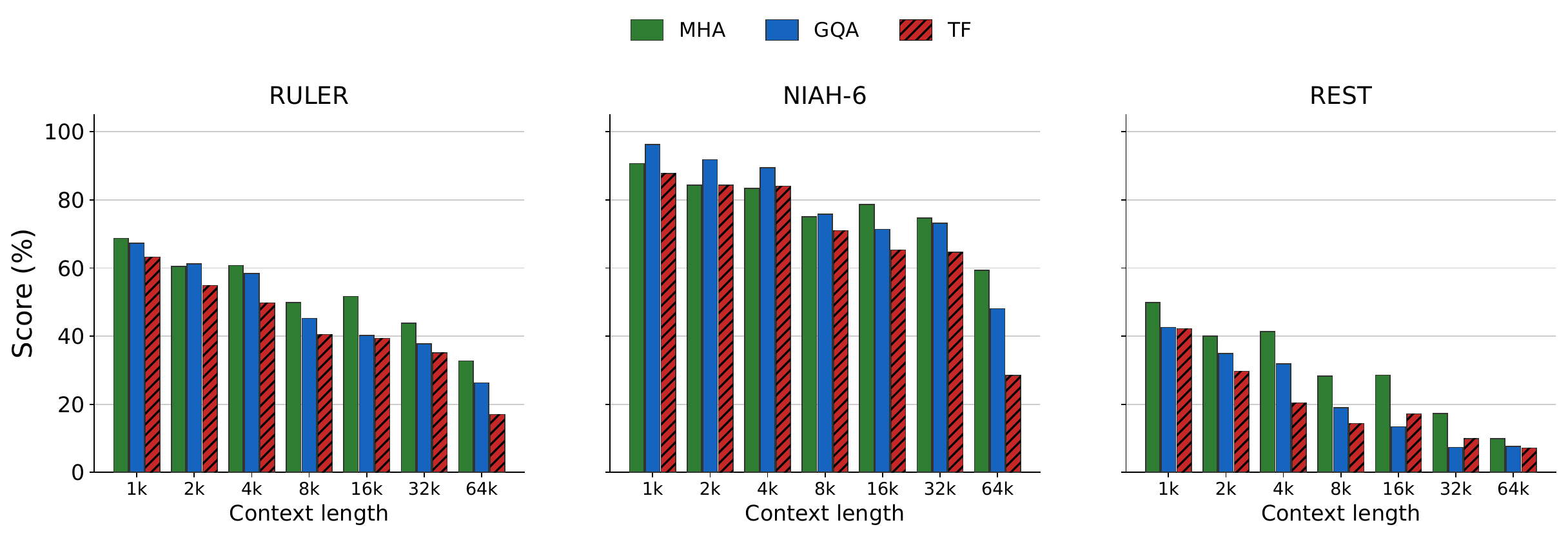}
  \caption{RULER, NIAH-6, and REST versus context length for the $+5$B checkpoints in Table~\ref{tab:gqa}.
  NIAH-6 is S1--S3, MK1, MV, and MQ; REST is the other seven tasks (MK2--3, VT, CWE, FWE, QA1--2).
   lengths 1k--64k.}
  \label{f:gqa_splits}
\end{figure*}

\begin{figure*}[h]
  \centering
  \includegraphics[width=\linewidth]{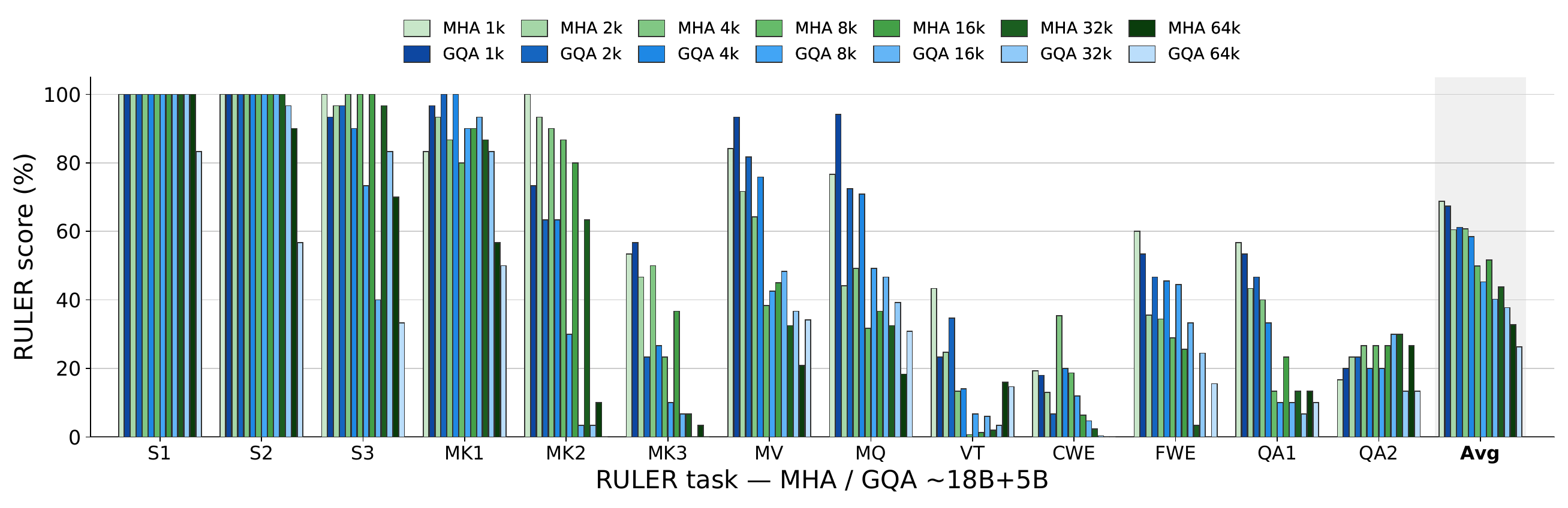}
  \caption{Per-task RULER scores for the 920M 18B ClimbMix $+5$B ProLong MHA and GQA checkpoints, at 1k / 2k / 4k / 8k / 16k / 32k / 64k.
  All 13 tasks are shown. Avg is the 13-task RULER mean, $n{=}30$.}
  \label{f:gqa_ruler}
\end{figure*}

Table \ref{tab:gqa} shows the performance of GQA on our 924M model, vs. our regular MHA Extender and MHA Transformer, at the 18B ClimbMix pretrain, +2.5B ProLong and +5B ProLong
checkpoints.
Because our 920M class model has $d_{model}=1664$, there are 13 heads each 128 features wide. To accomodate this, we use GQA4/5: two groups of 4 queries, one group of 5 queries.
Due to the GQA dimensionality reduction, the actual parameter count of the GQA4/5 model is 813M rather than 924M.

We find that GQA interacts well with Extender, and the Extender GQA4/5 model outperforms the Transformer MHA.
Due to resource constraints, we did not train a matching Transformer GQA.

Figure~\ref{f:gqa_core} provides per-task performance on the CORE benchmark. GQA tracks MHA performance closely on the more stable T-Mono tasks.
On the remaining tasks results are a little noisier, but GQA is largely tracking MHA there as well.
Figure~\ref{f:gqa_splits} shows the RULER performance divided into 13-task mean, NIAH-6, and REST.
GQA is overall tracking MHA closely on both the easier NIAH-6 tasks and REST.
Finally, Figure~\ref{f:gqa_ruler} shows individual task performance vs. context length.

\FloatBarrier
\clearpage
\section{Accuracy across model sizes}
\label{s:s0w0}

Table~\ref{tab:s0w0_sizes} reports centered CORE and RULER for the default Extender for our three size classes, for models trained on ClimbMix only, to a Chinchilla token budget.
Accuracy on both CORE and RULER 1--2k improves consistently with model size, particularly on T-mono, where the Transformer is also reliably improving with model size.
At context lengths outside of the 2k training context length, RULER scores are less predictable, but suggest an improvement as well.

\begin{table*}[h]
  \centering
  \small
  \setlength{\tabcolsep}{3.6pt}
  \caption{Default Extender at each size class in Figure~\ref{f:core}.
  199M is 4B tokens, 438M is 9.6B, and 924M is 18B.
  T-mono and CORE are centered DCLM CORE (\%); T-mono is the 12-task set in Figure~\ref{f:core}.
  RULER is the 13-task mean (\%).
  Italics are lengths longer than the 2k training context.
  ``--'': not evaluated.}
  \label{tab:s0w0_sizes}
  \begin{tabular}{lcccccc}
    \toprule
    Model & T-mono & CORE & \multicolumn{4}{c}{RULER} \\
    \cmidrule(lr){4-7}
     & & & 1k & 2k & 4k & 8k  \\
    \midrule
    199M & 20.4 & 15.7 & 49.3 & 41.0 & \textit{28.0} & \textit{16.9} \\
    438M & 26.0 & 19.5 & 57.3 & 45.1 & \textit{33.4} & \textit{15.6} \\
    924M & 32.7 & 23.7 & 59.0 & 50.3 & \textit{32.9} & \textit{19.9} \\
    \bottomrule
  \end{tabular}
\end{table*}

\end{document}